\documentclass[sigconf]{acmart}
\usepackage{amsmath}

\usepackage{amssymb}
\usepackage{caption}
\usepackage{epsfig}
\usepackage{graphicx}
\usepackage{booktabs}
\usepackage{multirow}
\usepackage{makecell}
\usepackage[misc,geometry]{ifsym}
\usepackage{siunitx}
\usepackage{color}
\usepackage{xcolor}
\usepackage{soul}
\usepackage{xcolor}
\usepackage{subfig}
\definecolor{OliveGreen}{rgb}{0,0.6,0}
\definecolor{SoftRed}{rgb}{1,0.2,0.2}
\usepackage[hyperpageref]{backref}
\usepackage{pifont}
\newcommand{\cmark}{\ding{51}}%
\newcommand{\xmark}{\ding{55}}
\AtBeginDocument{%
  }

\copyrightyear{2024}
\acmYear{2024}
\setcopyright{acmlicensed}
\acmConference[CODS-COMAD Dec '24]{8th International Conference on Data Science and Management of Data (12th ACM IKDD CODS and 30th COMAD)}{December 18--21, 2024}{Jodhpur, India}
\acmBooktitle{8th International Conference on Data Science and Management of Data (12th ACM IKDD CODS and 30th COMAD) (CODS-COMAD Dec '24), December 18--21, 2024, Jodhpur, India}
\acmDOI{XXXXXXX.XXXXXXX}

\begin{document}

\title{TabSniper: Towards Accurate Table Detection \& Structure Recognition for Bank Statements}

\author{Abhishek Trivedi}
\authornote{Both authors contributed equally to this research.}
\authornote{For queries, mail at Abhishek.Trivedi@aexp.com or Himanshu.S.Bhatt@aexp.com}
\affiliation{%
  \institution{American Express}
  \city{Bangalore}
  \state{KA}
  \country{India}}

\author{Sourajit Mukherjee}
\authornotemark[1]
\affiliation{%
  \institution{American Express}
  \city{Bangalore}
  \state{KA}
  \country{India}}


\author{Rajat Kumar Singh}
\affiliation{%
  \institution{American Express}
  \city{Bangalore}
    \state{KA}
  \country{India}}

\author{Vani Agarwal}
\affiliation{%
  \institution{American Express}
  \city{Bangalore}
    \state{KA}
  \country{India}}

\author{Sriranjani Ramakrishnan}
\affiliation{%
  \institution{American Express}
  \city{Bangalore}
    \state{KA}
  \country{India}}

\author{Himanshu Sharad Bhatt}
\authornotemark[2]
\affiliation{%
  \institution{American Express}
  \city{Bangalore}
    \state{KA}
  \country{India}}

\renewcommand{\shortauthors}{Trivedi et al.}

\begin{abstract}
Extraction of transaction information from bank statements is required to assess one’s financial well-being for credit rating and underwriting decisions. Unlike other financial documents such as tax forms or financial statements, extracting the transaction descriptions from bank statements can provide a comprehensive and recent view into the cash flows and spending patterns. With multiple variations in layout and templates across several banks, extracting transactional level information from different table categories is an arduous task. Existing table structure recognition approaches produce sub optimal results for long, complex tables and are unable to capture all transactions accurately. This paper proposes TabSniper, a novel approach for efficient table detection, categorization and structure recognition from bank statements. The pipeline starts with detecting and categorizing tables of interest from the bank statements. The extracted table regions are then processed by the table structure recognition model followed by a post-processing module to transform the transactional data into a structured and standardised format. The detection and structure recognition architectures are based on DETR\cite{carion2020endtoend}, fine-tuned with diverse bank statements along with additional feature enhancements. Results on challenging datasets demonstrate that TabSniper outperforms strong baselines and produces high-quality extraction of transaction information from bank and other financial documents across multiple layouts and templates.
\end{abstract}

\ccsdesc[500]{Computing methodologies~Computer vision problems}



\maketitle
\keywords{Table Detection\and Table Structure Recognition\and Finance\and Bank\and Detection Transformer\and Document\and Information Extraction \and Dataset}



\section{Introduction}
\begin{figure}[t]
    \centering
    \includegraphics[width=0.9\linewidth]{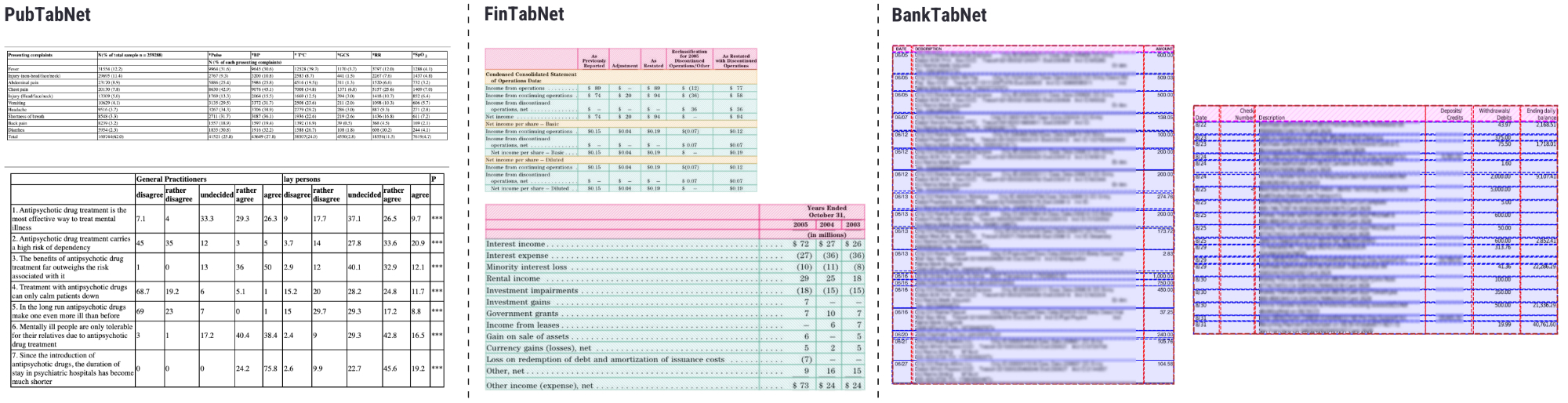}
    \caption{BankTabNet table images are relatively long and contain densely packed multi-line rows with intra-cell text spacing (misleads precise column detection) in comparison to publicly available Table Datasets.}
    \label{fig:Challenging_tables}
\end{figure}

Credit underwriting is a process of taking personal, financial, and/or business information, analyzing it for one’s ability to repay loans or other debts, and basis this accepting or rejecting the application. It could be for a new loan application or for a request to increase the credit limit. Bank statements are becoming more and more relevant for financial organizations to assess the customers’ financial well-being for credit ratings and underwriting. 
However, processing bank statements to extract the intended data in a structured format is not a trivial problem. While table structure recognition is a well-studied problem in literature; existing approaches fail to generalize well to Bank statement spreading. Processing banks statements at scale has several challenges in terms of variations in the layouts and templates across different banks, scanned v/s electronically generated PDF bank statements, variations in the representation of information and tabular structure. Tables in bank statements present specific information such as withdrawal, deposits, check etc. and hence tables needs to be categorized so that the extracted information can be processed appropriately. As shown in BankTabNet section of Fig \ref{fig:Challenging_tables}, tables in bank statements are comparatively more complex as they are longer tables spilling over multiple pages and comprise densely packed multi-line rows or wide text spacing in the columns. We believe, the problem of precisely extracting transactions from bank statements is unexplored and has several unique nuances. 

This paper presents first-of-its-kind end-to-end automated algorithm from bank statement spreading that extracts the transactional information from bank statements across multiple banks into a structured format. It starts with detecting and categorizing the tables in the bank statements. These detected table regions are then passed as inputs to the table structure recognition model which identifies the column and rows boundaries to detect multiple table cells. Finally, the paper presents a post-processing technique to extract the text corresponding to the detected table cells and transforming the output into a pre-defined standardised format. The post-spreading checks to tally the opening and closing balance with the running transaction balance makes our approach more deterministic. The key contributions of the paper are summarised below.
\begin{enumerate}
\item TabSniper- A first-of-its-kind end-to-end framework for table structure recognition for bank statements.
\item A table categorization technique to classify bank tables into different pre-defined classes.
\item A precise table structure recognition approach with innovative data preparation \& modelling techniques to cater to bank statement specific nuances such as handling long tables \& overlapping table objects. 
\item BankTabNet- An annotated database specific to the problem of table structure recognition from bank statements comprising annotations at both page and transaction level.
\end{enumerate}
\section{Related Works}
\label{sec:Related_work}
Tables in document images contains visually rich structured information.
With advancement in document AI space, end-to-end table structure recognition models play important role in industry applications. The challenges with table structure in document images is two fold: 1) variation in shape and size and 2) the noisy image background. Tables have complex structures, headers, empty or spanning cells, large/small blank spaces between neighboring columns, tables spanning across pages with multiple objects in between like text, paragraph etc. Although the problem of  table detection and structure recognition is well studied on controlled datasets, the above challenges limits the generalization of the existing approaches to other domains like healthcare, retail, banking etc.
Next, the paper summarizes the key research in the area of table detection and structure recognition.

\noindent \textbf{Table Detection}
Table detection problem was first explored with deep learning methods using Faster R-CNN based model  \cite{gilani2017table}. Other approaches including two-stage detectors viz Mask R-CNN  \cite{zheng2021global}, Cascade R-CNN \cite{cai2017cascade} and one stage detectors viz YOLO \cite{schreiber2017deepdesrt} were also used for table detection. Multiple image transformations like distance transform, augmentations, coloration etc. were attempted to improve the accuracy. Towards the unified detection head framework, dynamic head \cite{Dai_2021_CVPR} encompassing scale-awareness, task-awareness and spatial awareness was introduced. On the other hand, there are works using Non-maximum Suppression (NMS) to reduce the redundant predictions from the detector \cite{sun2021sparse,zhang2023dense}. DETR  \cite{carion2020endtoend} is an end-to-end object detector using set prediction loss with one-to-one assignments to predict the classes. Sparse R-CNN \cite{sun2021sparse} uses sparse learnable region proposals with dynamic interactive head to classify the objects.
 
\noindent\textbf{Table Structure Recognition}
covers row/column extraction methods which leverage object detection or semantic segmentation to detect the rows and columns, then merge them to form cells. DeepDeSRT \cite{schreiber2017deepdesrt} and TableNet \cite{paliwal2019tablenet} applied FCN based semantic segmentation to TSR. 
LGPMA \cite{qiao2022lgpma} uses soft pyramid masks as local and global levels to detect cell boundaries. PubTables-1M \cite{smock2021pubtables1m} defines separate components like table, row, column, spanning cell etc. to provide more structure to the table while detection, later merging the objects to form the grid cells.
TabStructNet \cite{raja2020table} and CascadeTabNet \cite{prasad2020cascadetabnet} provide end-to-end solution by combining the table element detection and cell prediction into a single network. 

Building a generalized solution across huge volume of bank statements with variety of templates is a non-trivial problem. Banking process automation helps to improve efficiency and reduce manual errors. In this digital era, information extraction for various banking documents like identification documents, account statements, investment statements, invoices, tax documents, financial reports etc. is becoming more and more relevant.  Most of the research and solutions in NLP and multi-modal document analytics \cite{gerling2023multimodal} are focused on document information extraction and on specific common documents viz financial statements  \cite{lewis2019fad}, banking orders  \cite{oral2019extracting} and others  \cite{baviskar2021efficient}. However, research on using deep learning solutions in banking domain remains limited, often focusing on specific document types. This gives a great opportunity for research in banking document analytics. To bridge the gap, this paper proposes TabSniper, an end-to-end table structure recognition method for bank statements. To the best of our knowledge, this is the first system that leverages state-of-the-art vision based model to extract bank transactions in a structured manner. 
\section{BankTabNet Dataset}
\label{sec:TDC_dataset}
For the bank transaction spreading task we have two separate models; first model is for Table Detection and Categorization (TDC) and the second model is for Table Structure Recognition (TSR). The publicly available datasets (Refer Fig \ref{fig:Challenging_tables}) were not specifically based on bank statement data and thus were not suitable for our requirements. Therefore, we created and annotated two datasets, for fine-tuning each of our models, using our in-house annotation team. Each page of bank statement pdf is initially converted to a RGB image. We have an in-house service capability to detect all Personally Identifiable Information data on pdf page images. Details like customer name, customer address, account number and account type are detected as key-value pairs and their respective bounding boxes are then masked. Thereafter, the TDC dataset is built on page images and TSR dataset is built by cropping the table images on each of the page image.
\subsection{TDC Dataset}
\label{sec:TDC_dataset_1}
For the TDC task we have two major objectives; one is to properly identify and categorize each and every table in the bank statement and second is to identify/refine the categories of the tables with the help of table captions and headers. Based on these objectives, we annotated the bank statement tables into the categories explained in Table \ref{tab:tdc_categories} and created a vision-based and a text-based dataset.
\subsubsection{Vision-based Dataset:}
We annotated around $11,607$ page images for different classes from bank statements. Krippendorff's Alpha (K-alpha)\cite{kalpha} score is used for calculating the Inter-Annotator Agreement (IAA) and ensure minimum annotation bias. K-alpha utilizes the Intersection over Union (IoU) value of the annotated bounding boxes across the different annotators for evaluating the agreement between them. K-alpha score range lies between $0$ (no-agreement)  and $1$ (complete agreement). 
Based on this scoring mechanism the calculated average IAA score across all the samples for $IOU>0.5$ is $0.955$ and for $IOU>0.9$ it is $0.994$. Both of these scores imply high agreement among annotators. The TDC vision based dataset has $7544$ bank statement training images, $2322$ test images and $1741$ validation images.
\begin{table}[!htbp]
\caption{TDC table categories}\label{tab:tdc_categories}
\centering
\resizebox{0.47\textwidth}{!}
{
\begin{tabular}{|l|l|l|}
\hline
Table Category & Count & Description\\
\hline
Credit & $2742$ &Tables having caption related to the keyword "credit".\\
Debit & $3202$ & Tables having caption related to the keyword "debit". \\
Check & $3294$ & Check transaction related tables \\
Txn\_bal & $2106$ & Transaction tables having both credit and debit columns. \\
Txn\_amt\_bal & $965$ & Transaction tables having an amount column where negative.\\
& & amount implies "debit" and positive amount implies "credit".\\
Txn\_check\_bal & $401$ & Transaction tables having a column related to check number. \\
Summary\_accounts & $715$ & Tables containing summary of multiple accounts\\
Statement\_summary & $2013$ & Tables containing summary of a single account \\
Service\_fees & $639$ & Tables containing details about service fees.\\
Check\_image & $20313$ & Check images in the bank statement. \\
Table\_caption & $16519$ & Includes all table captions\\
Table\_header & $20313$ & Includes all table column headers\\
\hline
\end{tabular}
}
\end{table}
\subsubsection{Text-based Dataset:} \label{sec:TDC_text_dataset}
For refining the categories detected by DETR \cite{carion2020endtoend}, we create another text based dataset using the table captions and headers from the annotated dataset. To create this dataset, we capture the text content in the table caption and header regions by finding the OCR bounding boxes having the highest overlap i.e. $IOU > 0.5$ with the annotated caption and header bounding boxes. For each image, we generate the OCR bounding boxes and related text using off-the-shelf OCR service. The next step involves mapping the captions and headers to their respective tables. Since there can be multiple captions in a page so we map each caption to its associated table based on the minimum vertical distance of their top left y-coordinates.
In almost all the cases the header lies completely within the table boundaries. Hence for mapping headers to tables in a certain page, we calculate the ratio of intersection area of the header and table bounding boxes to the header bounding box area. If this ratio is equal to or close to 1 then it indicates that the header lies almost within the table and we map the header to that particular table. After the mapping process is complete, we use the annotated table labels to create the dataset for training our text based classifier. Since several tables miss having a header or caption, we  
split the dataset into three separate categories: "header\_only" (total $9321$ samples), "caption\_only" (total $9792$ samples) and "both\_header\_caption" (total $7868$ samples). Using these $3$ datasets we trained $3$ separate text-classifiers as explained in Section \ref{sec:tdc_desc}.

\subsection{TSR Dataset}
\label{sec:TSR_dataset}
For TSR task, the objective is to identify the structure of the table and mark the bounding boxes for all the categories defined in Table \ref{tab_tsr_categories}. We annotated around $5165$ tables from $310$ bank statements. These tables were from different categories mentioned in Table \ref{tab:tdc_categories}. 
Since the TSR annotation process is also related to object detection so we have used K-alpha \cite{kalpha} for calculating IAA. Approach described in Section 3.1 is calculating the IAA score for TSR dataset. Based on that scoring mechanism the calculated average IAA score across all the samples for $IOU > 0.5$ is $0.99$ and for $IOU > 0.9$ it is $0.98$. Both of these scores imply high agreement among annotators.
\begin{table}[!htbp]
\caption{TSR table categories}\label{tab_tsr_categories}
\centering
\resizebox{0.47\textwidth}{!}
{
\begin{tabular}{|l|l|l|}
\hline
Table Category & Count & Description\\
\hline
Table & 5165 & Table bounding box for any tabular data identified in a bank statement. \\
Table Row & 53178 & Bounding boxes for each horizontal row inside a table. \\
Table Column & 18448 & Bounding boxes for each vertical column inside a table. \\
Table Column Header & 4569 & If first row of the table have a header or description of details about column\\ 
& & values then mark it as table column header. \\
Table Spanning Row & 912 & If any row spans across multiple columns then mark it as table spanning row.\\
\hline
\end{tabular}
}
\end{table}

\noindent \textbf{Dataset Preparation:} We further introduce different kinds of padding of white pixels ($20$ and $40$ pixels) around annotated table images to make the model robust to different sizes of table images extracted from the TDC model. When marking the bounding boxes for consecutive rows, we made sure that there is minimal or no overlap between the bounding box boundaries.  
Different ablation studies (Table \ref{tab_tsr_result1}) are designed with respect to these versions of dataset to increase the efficacy of our algorithm. The table images go through initial pre-processing stages of long tables' split, padding variations addition, normalization before setting up the Data Loader of TSR model. 
After these data ablation experiments, we have $9724$ training images, $2000$ validation images and $2200$ test images for TSR dataset.
\begin{figure}[!htbp]
    \centering
    \includegraphics[width=0.85\linewidth]{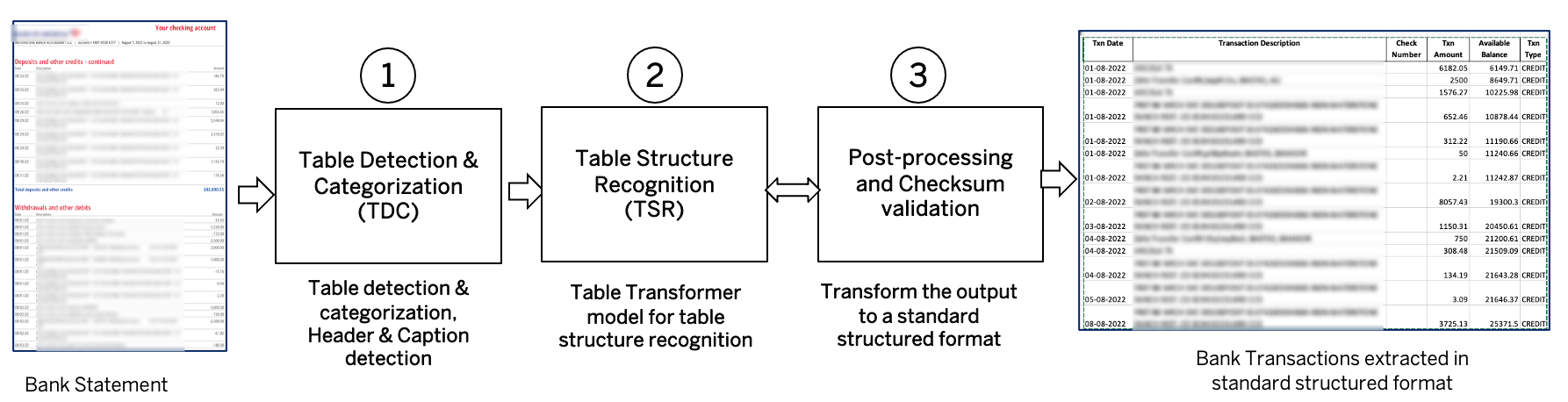}
    \caption{Overall flow of TabSniper}
    \label{fig:overall_flow}
\end{figure}
\section{TabSniper Model}
\label{sec:tabsniper}
\noindent \textbf{Overview:} Given a bank statement, the objective is to extract all transactions (e.g. credit, debit, check number) present in multiple tables across different pages. The pipeline consists of three different stages -- see Fig.~\ref{fig:overall_flow}. In the first stage, page images of bank statements are processed by Table Detection \& Categorization model to obtain bounding box and table categorization (10 classes). The raw bank statements are also fed to off-the-shelf OCR service which returns the text and its bounding box co-ordinates. The table image along with the OCR text are fed as input to the Table Structure Recognition model. TSR predicts bounding boxes and labels associated with  different object classes forming table structure for every table image in pipeline. Once we identify the row, column, column header, and spanning cells, post-processing is performed based on the heuristics to adjust the position of these objects to reduce the overlap and provide structure of the table. The intersection of each pair of table column and table row objects can be considered to form a separate implicit class, table grid cell. The text inside the table cell is matched with OCR word position and returns the corresponding text of the table cell. A tabular data frame is constructed from text information in each cell for respective table images. The data frames are then passed through final post processing stage where each table is sequentially processed to extract all the transaction level information 
Next, we describe the components of TabSniper along with a brief about DETR \cite{carion2020endtoend}.
\begin{figure*}[!htbp]
    \centering
    \includegraphics[width=0.9\linewidth]{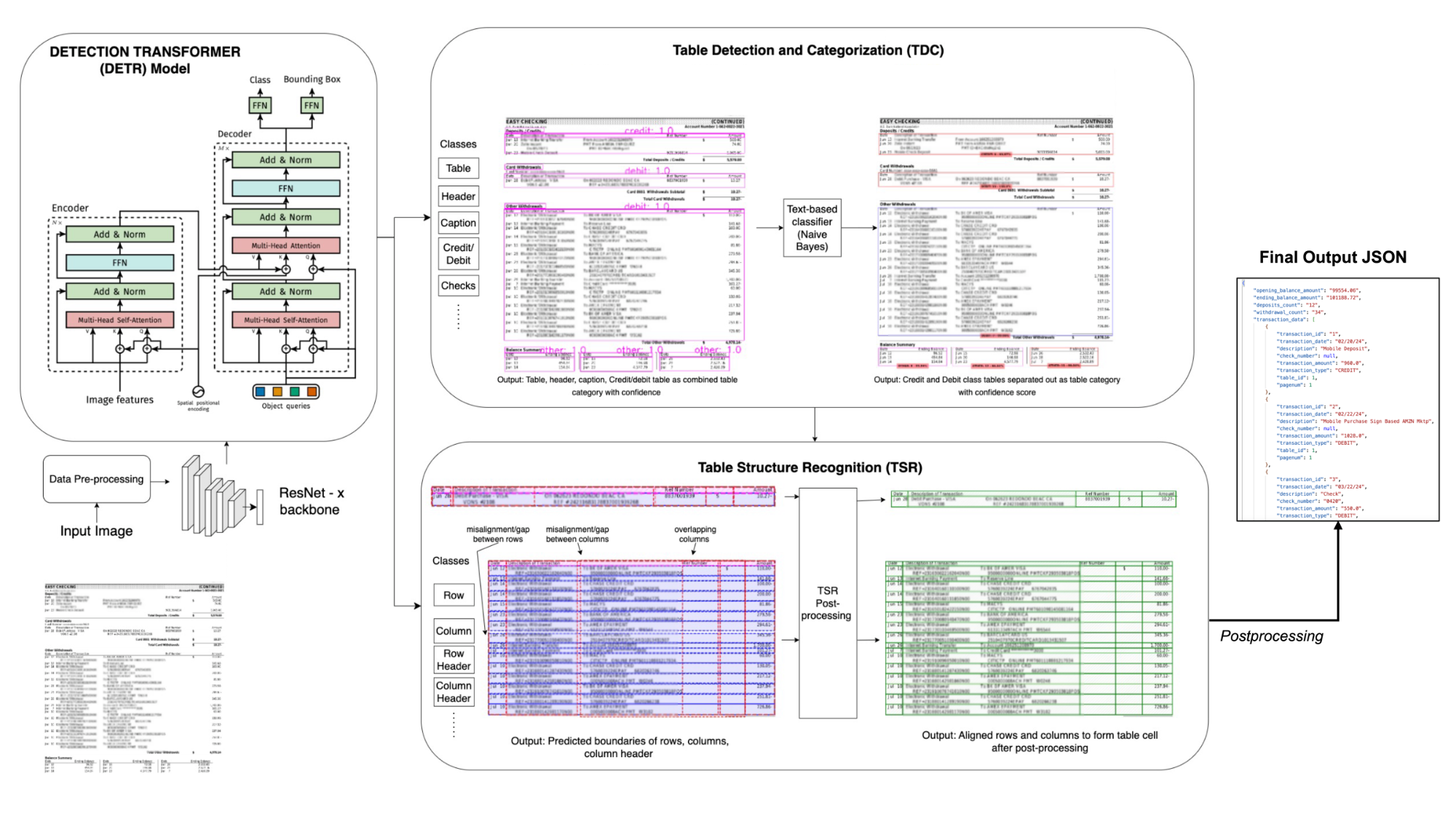}
    \caption{Architecture diagram of TabSniper}
    \label{fig:archi}
\end{figure*}
\subsection{Detection Transformer}
In the current implementation of TabSniper, we use Detection Transformer (DETR) \cite{carion2020endtoend} for model building. A transfer learning technique is used to initiate the transformer weights with further fine-tuning on BankTabNet(Section \ref{sec:TDC_dataset}). DETR \cite{carion2020endtoend} is pre-trained on the PubTables-1M, which contains nearly one million tables from scientific articles, supports multiple input modalities, and contains detailed header and location information for table structures, making it useful for a wide variety of modelling approaches. 
The key novelty in DETR model is to predicts all objects at once, and is trained end-to-end with a set loss function which performs bipartite matching between predicted and ground-truth objects. 

\subsection{TDC - Table Detection \& Categorization}
\label{sec:tdc_desc}
\begin{figure}[!htbp]
    \centering
    \includegraphics[width=0.9\linewidth]{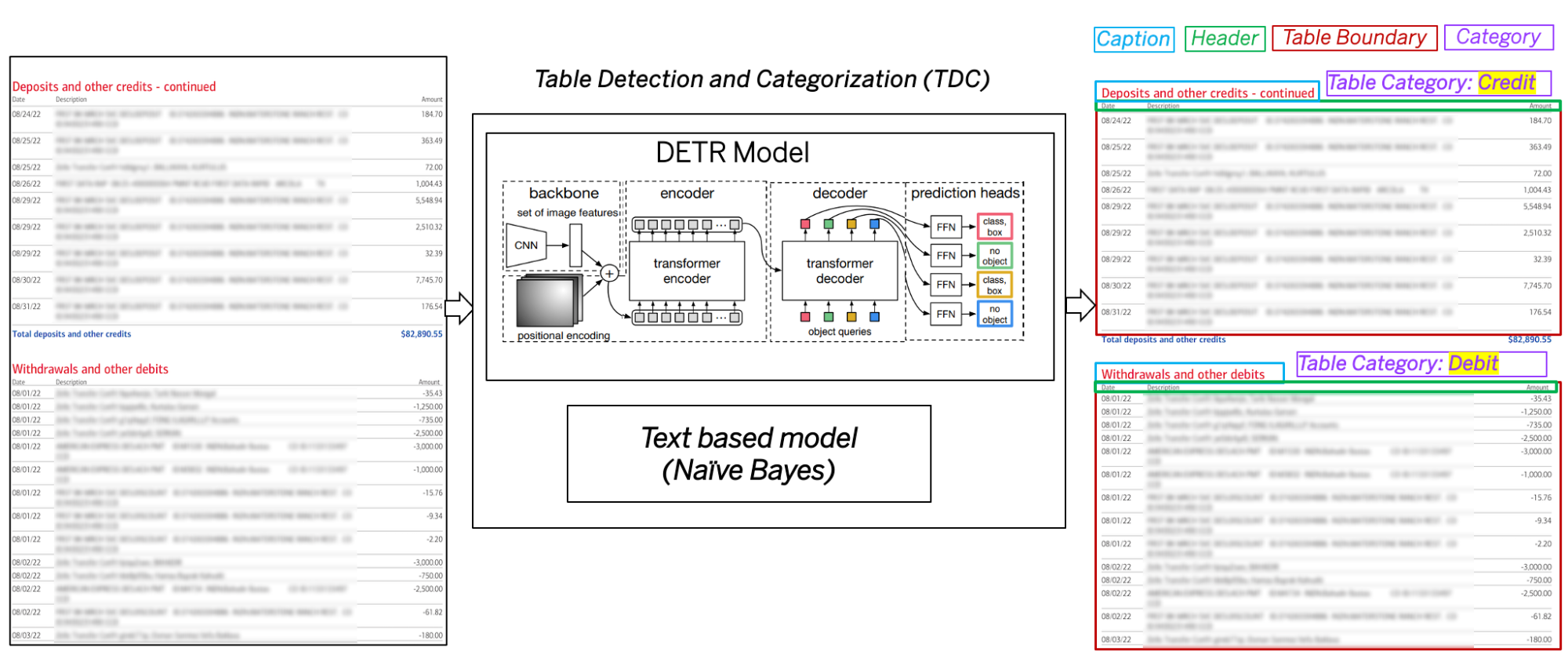}
    \caption{Table Detection and Categorization flow}
    \label{fig:TDC_flow}
\end{figure}

For TDC, we followed a sequential multi-modal approach using both image and text. It starts by passing the bank page image through a vision based model for detecting and categorizing the tables. Post that we pass the extracted table captions and headers through a text based classifier for segregating and refining the table categories (see Fig. \ref{fig:TDC_flow}).

\noindent \textbf{Vision-based Model:}\label{sec:tdc_vision_desc} For the vision model, along with DETR \cite{carion2020endtoend}, we also tried other popular baselines shown in Table \ref{tab_tdc_result1}. We start with all the table categories mentioned in Table \ref{tab:tdc_categories}.
However, since caption was not part of table boundary and it is the only thing that differentiates credit and debit tables, we combined credit and debit as single table category Credit/Debit. This class gets eventually refined via Text-Based model. We also combined Summary\_accounts, Statement\_summary, Service\_fees as single table category named Other. Signed check images in bank statements form separate class to avoid mis-classification as table image. 
DETR \cite{carion2020endtoend} is trained using three different kinds of losses. The Cross-Entropy(${L}_{CE}$) loss targets accurate class prediction while L1(${L}_{l1}$) and Generalized IoU loss(${L}_{GIoU}$) optimizes bounding box detection. Next, we state loss functions along with a brief description about the GIoU loss.
\paragraph{Generalized Intersection over Union (GIoU) :} 
GIoU loss is a modified version of IoU loss function that has an addition penalty term.
\begin{align}
{L}_{GIoU} = 1 - \frac{|(B \bigcap {B}_{gt})|}{|(B \bigcup {B}_{gt})|} + \frac{|C-B\bigcup{B}_{gt}|}{|C|}
\end{align}
where C is the smallest box covering B(predicted bounding box) and ${B}_{gt}$ (ground truth bounding box). 
Due to the introduction of the penalty term, GIoU loss keeps expanding the size of the predicted box until it overlaps with the target box, and then the basic IoU term will work to maximize the overlap area of the bounding box. The final loss equation for DETR \cite{carion2020endtoend} becomes-
\begin{align}
{L}_{DETR} = \sum_{n=1}^{N} {\lambda}_{ce}.{L}_{CE} + {\lambda}_{l1}.{L}_{L1} + {\lambda}_{giou}.{L}_{GIoU}
\end{align}
DETR \cite{carion2020endtoend} infers a fixed-size set of $N$ (number of queries) predictions, in a single pass through the decoder. ${L}_{l1}$ is normal box regression loss set as $|B-{B}_{gt}|$. Weight coefficients of individual losses ${\lambda}_{ce}, {\lambda}_{l1} \& {\lambda}_{giou}$ form model hyper-parameters.
\\

\noindent \textbf{Text-based Model:}\label{sec:tdc_text_dataset} The output of the table detection and categorization is then passed to the text-based classifier for category segregation and rectification. Table structures for Debit and Credit category were almost similar. To solve this problem, we use the $3$ TDC textual datasets (described in Section \ref{sec:TDC_text_dataset}) for training $3$ separate Multinomial Naive Bayes (NB) Classifiers  \cite{manning-2008} : Header NB, Caption NB and Header\_Caption NB classifier models. The three models are based on the availability of table caption and header for a particular table. Here, we chose to use a multinomial NB Classifier as it
takes into consideration the count of words in a certain category, which is useful when some words repeat in a certain header or caption for a particular category. For the TDC text classifier model (refining categories predicted by DETR) the Header\_Caption NB model performs the best across all the different categories(Table \ref{tab:tdc_nb_result}).
\begin{table}[!htbp]
\centering 
\caption{Text Classifier F1-Scores
on TDC text-dataset}\label{tab:tdc_nb_result}
\resizebox{0.47\textwidth}{!}
{
\begin{tabular}{|l|l|l|l|l|l|l|l|}
\hline
Model &  Credit & Debit & Check & Txn\_bal & Txn\_amt\_bal & Txn\_chk\_bal & Other\\
\hline
Header NB & 0.48& 0.63& 0.77 & 0.90 & 0.78 & 0.87 & 0.74 \\
Caption NB & 0.93 & 0.93 & 0.95 & 0.77 & 0.63 & 0.86 & 0.84 \\
Header\_Caption NB & \textbf{0.95} & \textbf{0.94} & \textbf{0.96} & \textbf{0.92} & \textbf{0.89} & \textbf{0.94} & \textbf{0.89} \\
\hline
\end{tabular}
}
\end{table}
\subsection{TSR - Table Structure Recognition}
\label{sec:tsr_post}
\begin{figure}[!htbp]
    \centering
    \includegraphics[width=0.9\linewidth]{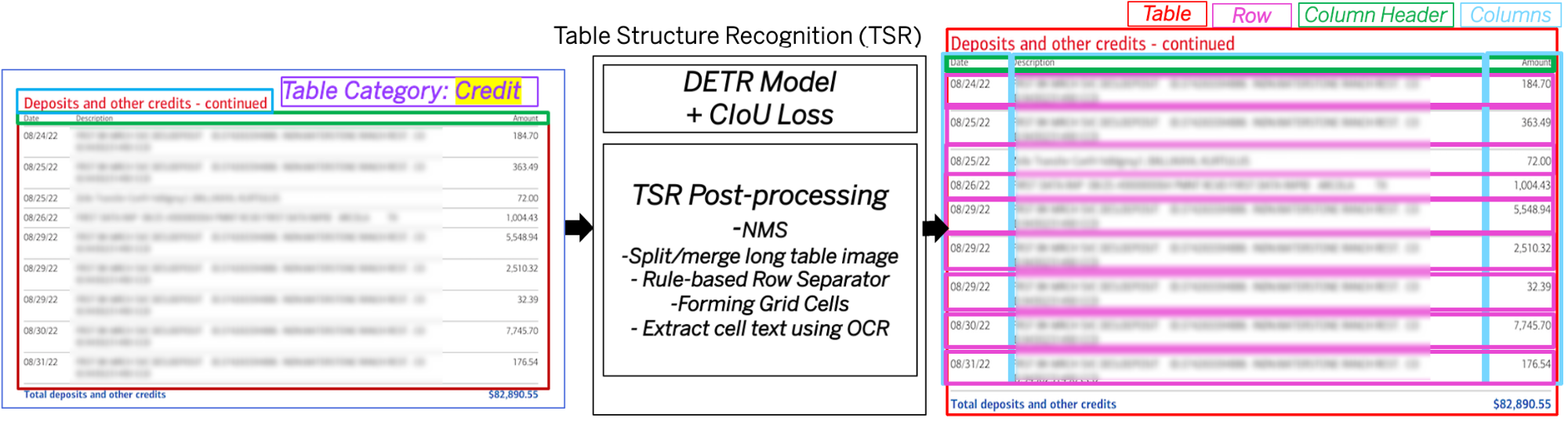}
    \caption{TSR flow diagram}
    \label{fig:TSR_flow}
\end{figure}

Using the above DETR \cite{carion2020endtoend} model as base, the next step is to perform table structure recognition (TSR). We use a novel approach to model TSR using five object classes: table, table row, table column, table column header and table spanning cell (shown in Fig \ref{fig:archi}). These objects model a table’s hierarchical structure through physical overlap. The raw bank statements are fed to off-the-shelf OCR service which returns the text and their co-ordinates. The table image along with the OCR text are fed as input to the TSR model. The number of queries ($N$) parameter of the transformer is set to $125$ to accommodate both inference speed and precision. With the above static $N$ parameter, the model detects different classes (Section \ref{sec:TSR_dataset}) inside short and medium sized table efficiently. Long tables(>20 rows) are split horizontally into two sub-images before passing to TSR model. We track the split images and merge them at post-processing level. This ensures effective processing with optimum inference speed for all lengths of table. The intersection of each pair of table column and table row objects can be considered to form a seventh implicit class, table grid cell. The text inside the table cell is matched with OCR word position and returns the corresponding text of the table cell. The model flow is shown in Fig \ref{fig:TSR_flow}.

\paragraph{Training TSR:} 
We modified Generalized IoU loss (GIoU) of basic DETR \cite{carion2020endtoend} to Complete IoU loss (CIoU) in our final TSR model to address the inaccurate mismatch of table objects with ground truth. The vertically oriented dense table rows in bank statement tables are often matched with incorrect GT rows during model training due to non-involvement of several other geometric factors like overlapping area, distance, and aspect ratio in Generalized IoU Loss function. This results in excess false positive bounding box row candidates which are difficult to filter through existing NMS techniques. Complete IoU loss function incorporates all the the above geometric factors, enhances speed and accuracy all at once.

\paragraph{Distance Intersection over Union (DIoU) :} 
DIoU loss incorporates the normalized distance between the predicted box and the target box along with basic GIoU penalty term. It inherits properties from both IoU and GIoU and further improves the table spreading by directly minimizing the distance between two boxes and thus converges much faster than GIoU loss.  
\begin{align}
{L}_{DIoU} = 1 - \frac{|(B \bigcap {B}_{gt})|}{|(B \bigcup {B}_{gt})|} + {R}_{DIoU}(B, {B}_{gt})
\end{align}
\begin{align}
    {R}_{DIoU}  =  \frac{{p}^{2}(B,{B}_{gt})}{{c}^{2}}
\end{align}
\noindent The penalty term ${R}_{DIoU}$ is Euclidean distance between center coordinates of the target bounding box and prediction bounding box normalized by $c$ the diagonal length of the smallest enclosing box covering both prediction \& GT bounding boxes. It is a scale-invariant function and provides moving directions for predicted bounding boxes towards GT target box.
\paragraph{Complete Intersection over Union (CIoU) :}
The CIoU Loss is the modified version of DIoU loss function. Complete IoU loss takes one step ahead and also accommodates deviation of aspect ratio over DIoU loss function.
\begin{align}
    {L}_{CIoU} = {L}_{DIoU} + {\alpha * \upsilon}
\end{align}
where,
\begin{align}
    {\upsilon} = \frac{4}{{\pi}^2}{(\arctan\frac{{w}^{gt}}{{h}^{gt}} - \arctan\frac{w}{h})}^{2}
\end{align}
$\alpha$ is a trade-off parameter and is defined as
\begin{align}
    {\alpha} = \frac{\upsilon}{(1-IoU) + \upsilon}
\end{align}

\noindent The final modified loss equation for TSR built on DETR \cite{carion2020endtoend} with weight coefficients  $\lambda$ is -
\begin{align}
{L}_{TSR} = \sum_{n=1}^{N} {\lambda}_{ce}.{L}_{CE} + {\lambda}_{l1}.{L}_{L1} + {\lambda}_{ciou}.{L}_{CIoU}
\end{align}
\paragraph{TSR PostProcessing: }
In our entire pipeline we call the OCR service only once at the beginning, across the different pages. Hence after the TDC output gets generated, we are able to generate the rule-based row-separation bounding boxes using the top aligned OCR words in left most Date column of table images.
These row-separator outputs assists the TSR post-processing to refine row predictions. It is especially useful in detecting rows for long tables or tables with very densely packed rows. Post-processing is also performed based on the heuristics to adjust the position of aforementioned six classes to reduce the intra-class overlap and give the structure of the table. These heuristics include NMS, filling row/column gaps and addition of missing row and columns to ensure complete and precise detection of each transaction.
\subsection{PosP - Postprocessing}
After obtaining each table category from the TDC module and the table contents from the TSR module, the final step of the pipeline is to extract all the transactions appropriately. Primarily, for every extracted transaction we need to decide whether it will be of ``Debit" category or ``Credit" category. Also to extract the transactions in the order in which they appear in the bank statement we sort all the tables based on their page numbers, followed by sorting the tables within the same page based on their vertical y-coordinates. To extract transaction details from tables, synonyms for key words like ``Date", ``Description", ``Amount", ``Check Number", ``Credit", ``Debit" and ``Balance" are created and matched with table header words to compute the respective header indices based on which the transaction details are extracted. The table categories are used for deciding the category of a transaction . Transactions from ``Credit" and ``Debit" labelled tables are classified as ``Credit" and ``Debit" respectively. Transactions from ``Txn\_bal" and ``Txn\_check\_bal" labelled tables are marked based on whether their ``Credit" or ``Debit" columns are non-empty. For ``Txn\_amt\_bal" labelled tables we check presence of a minus (-) sign or enclosed brackets in the extracted amount to decide whether the transaction belongs to "Debit" category or ``Credit" category. For every extracted transaction we make sure that its  ``Date" and ``Amount" exists otherwise it is discarded. After extracting all the transactions, we compute the checksum value as follows: $Open_{Bal} - \sum_{t \in Debit}t + \sum_{t \in Credit}t - End_{Bal}$. Here $t$ refers to an extracted transaction. The opening and ending balances ($Open_{Bal}$ and $End_{Bal}$) for a particular bank statement are obtained using an in-house statement summary extraction service. For a particular statement if the checksum value is computed as $0$ then we consider that all the transactions were correctly extracted for that statement.
\section{Experimental Setup}
\label{sec:experiment}
\subsection{Implementation Details}
\subsubsection{TDC}
We have used Detectron2 \cite{wu2019detectron2} for implementation of TDC vision model DETR \cite{carion2020endtoend}. At training, scale augmentation for the shortest side of the page image is kept in range of $400-800$ pixels, while keeping the aspect ratio unchanged. 
Hyper-parameters for TDC vision model are set to following after tuning on validation set: Learning Rate (LR): $1e-5$, Number of Epochs: $100$,  Batch Size: $16$, ${\lambda}_{ce}$  (weight coefficient of cross entropy loss): 1, ${\lambda}_{l1}$ (weight coefficient of L1 loss): 5, and ${\lambda}_{giou}$ (weight coefficient of  Generalized IoU loss): 2. The mAP of the TDC vision model after training is shown in table \ref{tab_tdc_result1}.
\\
For the TDC text based model, we trained all three NB model using dataset split of $70\%$ train, $30\%$ test data. When developing the classifier we utilized a Count Vectorizer which takes into consideration the count of each word during vectorizing each sample of table caption or header. Total caption and header words vocabulary counts were $840$ and $860$ respectively. For tables containing both header and caption, we concatenated the caption and header words and passed to the classifier during training and inference. Although we include "Service\_Fees" category under "Other" (Section \ref{sec:tdc_vision_desc}) category due to low sample count (Table \ref{tab:tdc_categories}), however some service fees tables contains "Debit" related transactions. To extract these transactions, we perform a word check on the caption of "Other" class tables for the word "Service Fee" and mark them as "Debit" category during inference.
\subsubsection{TSR}
The input $H \times W \times 3$ table RGB image is processed by TSR to detect bounding box and class label associated with six different regions in table. A tabular dataframe is finally extracted through physical overlap of regions of interest. At training time, scale augmentation is used such that the longest side is randomly set in range of $1100-1300$ pixels range while keeping the aspect ratio intact. 
At inference time, the input image size is set to $1200*1200$ pixels. In addition, the text content and location is derived from off-the-shelf OCR to enable overlap with seventh implicit region of table i.e table cells.
Hyper-parameters for training regime include Learning Rate (LR) and Number of Epochs. The learning rate is set to $5e-5$ which drops linearly after every $4$ epochs and Number of epochs is set to $100$ respectively. Our DETR \cite{carion2020endtoend} based TSR model has $4$ more hyperparameters namely Batch Size, ${\lambda}_{ce}$ , ${\lambda}_{l1}$ and ${\lambda}_{ciou}$ (weight coefficient of Complete IoU loss). These hyperparameters are fixed and set to $2, 1, 5$ and $2$ respectively based on tuning through validation set. Adam Optimizer is used to converge the loss function and attain the good performance. 
We trained both TDC \& TSR modules on NVIDIA A100 Tensor core GPU 40GB GPUs. The end-to-end inference time for a bank statement pdf with average of 20 pages on a Apple M1 Max CPU with 32GB memory is around 1.91 minutes.
\subsection{Performance Metrics}
We report the Average Precision (${AP}_{50},{AP}_{75},AP$) and Average Recall (AR) scores at different IoU thresholds for the models on the BankTabNet dataset in table \ref{tab_tdc_result1} for TDC and table \ref{tab_tsr_result1} for TSR.
\noindent For all the ablation models and final version, we use performance on the validation set to determine the optimal hyperparameters and modelling choices. Subsequently, we optimize the models on the combined training and validation splits and conduct a one-time evaluation on the test split. For the TDC text classifier we have compared the performance of different variants at the categorical level using the F1-score metric in table \ref{tab:tdc_nb_result}.
\subsection{Baselines \& Ablations}
Extensive comparative evaluations are performed for TabSniper against different state-of-the-art approaches. This includes Faster R-CNN \cite{7485869}, Mask R-CNN \cite{8237584}, Cascade R-CNN \cite{cai2017cascade}, DiT \cite{li2022dit}, and Dynamic Head \cite{Dai_2021_CVPR} for TDC-Vision model \& Table Transformer  \cite{smock2021pubtables1m}, LGPMA  \cite{qiao2022lgpma}, TabStructNet  \cite{raja2020table} for TSR in table \ref{tab_tsr_result1}. 
\section{Results}
\label{sec:result}
\subsection{Quantitative Results}
TabSniper-TDC outperforms Faster R-CNN, Mask R-CNN, Cascade R-CNN and achieves comparable results with more complex models like DiT and Dynamic Head while having significantly lower inference time. Our use-case application requires user to upload bank statements in real time without any API calls to GPU based servers. Therefore, optimal CPU inference time is critical for our use-case and DETR \cite{carion2020endtoend} based TabSniper-TDC is best choice for it. Table \ref{tab_tdc_result1} shows the inference time on Apple M1 Max CPU with $32$GB memory.
\begin{table}
\centering 
\caption{Performance of baselines \& TabSniper on TDC dataset (refer Section \ref{sec:TDC_dataset_1})}
\label{tab_tdc_result1}
\resizebox{0.47\textwidth}{!}
{
\begin{tabular}{|l|l|l|l|l|l|l|l|l|l|}
\hline
Model & Backbone & $AP$ & $AP_{50}$ & $AP_{75}$ & \makecell{CPU\\Inference} & \makecell{CPU Inference\\Time (s/image)}\\
\hline
Faster R-CNN \cite{7485869} & ResNet-50 & 83.50 & 94.55 & 90.25& \cmark & 1.71\\
Mask R-CNN \cite{8237584} & ResNet-50 & 83.84 & 95.14 & 90.13 & \cmark & 1.72\\
Cascade R-CNN \cite{cai2017cascade} & ResNet-50 & 84.08 & 93.07 & 89.18 & \cmark & 2.34\\
\textbf{TabSniper- TDC} & ResNet-50 & 85.25 & 93.91 & 90.69 & \cmark & \textbf{1.25}\\
DiT-B(Cascade) \cite{li2022dit} & Transformer & 87.96 & 95.73 & 92.57 & \cmark & 8.85\\
Dynamic Head \cite{Dai_2021_CVPR} & Swin-Tiny & 89.04 & 97.39 & 94.43 & \xmark & -\\
\hline
\end{tabular}
}
\end{table} 
\\
TSR baselines are efficient in processing structure of short, medium tables but misses out on precise row, column detection for long tables. Error rate in long bank tables is usually high due to missing row bounding boxes and partial capture of long transaction description inside table columns. We experimented with different ablations (Table \ref{tab_tsr_result1}) of TabSniper-TSR to precisely detect every object forming the table structure.
\begin{table}[!htbp]
\centering 
\caption{Performance of baselines \& TabSniper on TSR dataset (refer Section \ref{sec:TSR_dataset})}
\label{tab_tsr_result1}
\resizebox{0.47\textwidth}{!}
{
\begin{tabular}{|l|l|l|l|l|l|}
\hline
Model & Ablation & $AP_{50}$ & $AP_{75}$ & $AP$ & $AR$ \\
\hline
TabStructNet \cite{raja2020table} & - & 79.2 & 71.0 & 65.4 & 73.6 \\
LGPMA \cite{qiao2022lgpma} & - & 84.8  & 76.1 & 70.0 & 78.8 \\
Table-Transformer \cite{smock2021pubtables1m} & DETR (Base) & 85.3 & 76.5 & 70.4 & 79.3 \\
\textbf{TabSniper- TSR} & + Split-Merge Long Tables & 92.6 & 80.8 & 72.5 & 81.2 \\
 & + Padding Variations & 94.6 & 89.8 & 80.2 & 87.2 \\
 & + Complete IoU Loss & \textbf{94.8} & \textbf{91.5} & \textbf{83.1} & \textbf{90.6} \\
\hline
\end{tabular}
}
\end{table}
\vspace{-1em}
\paragraph{External Datasets:}We also benchmark TabSniper-TSR on two major publicly available table structure recognition datasets. Tables in Pubtables 1M \cite{smock2021pubtables1m} are derived from scientific articles and tables in FinTabNet \cite{zheng2021global} are derived from financial documents respectively.
Table \ref{tab_tsr_dataset} showcases performance metrics comparison for both TabSniper-TSR and Table Transformer(current SOTA) models when trained specifically on each of the external datasets. TabSniper has comparable ${AP}_{50}$ \& ${AP}_{75}$ scores but significantly outperforms Table-Transformer on $AP$ \& $AR$ scores, demonstrating better performance in table objects localization over range of IoU thresholds.
\begin{table}[!htbp]
\centering 
\caption{Results on Test Set of Publicly Available TSR Datasets}
\label{tab_tsr_dataset}
\resizebox{0.47\textwidth}{!}
{
\begin{tabular}{|l|l|l|l|l|l|}
\hline
External Dataset & Model & $AP_{50}$ & $AP_{75}$ & $AP$ & $AR$ \\
\hline
Pubtables-1M \cite{smock2021pubtables1m} & Table-Transformer & 96.3 & 92.3 & 84.4 & 89.3 \\
 & TabSniper- TSR & 96.2  & 93.8 & 89.6 & 93.3 \\
\hline
FinTabNet \cite{zheng2021global} & Table-Transformer & 97.4 & 94.2 & 88.8 & 93.1 \\
& TabSniper- TSR & 96.0 & 93.0 & 89.1 & 93.4 \\
\hline
\end{tabular}
}
\end{table}
\subsection{Qualitative Results}
\subsubsection{TabSniper-TDC}
\begin{figure}[!htbp]
    \centering
    \includegraphics[width=0.9\linewidth]{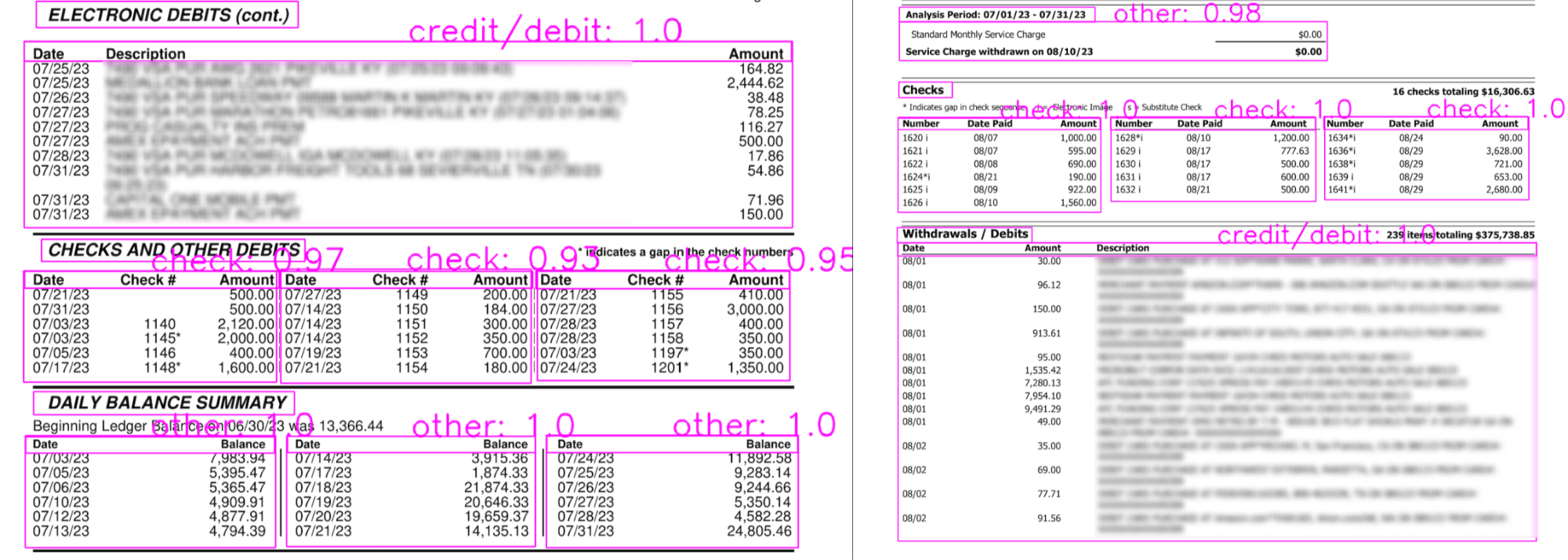}
    \caption{Table detection and categorization on different bank statements using TabSniper.}
    \label{fig:result_tdc}
\end{figure}

TDC qualitative results are comparable for full table detection and categorization. However, TabSniper- TDC takes less inference time with respect to all mentioned baselines in Table \ref{tab_tdc_result1}. Table bounding boxes with category prediction using TabSniper TDC vision model is shown in Figure \ref{fig:result_tdc}.\\
\subsubsection{TabSniper-TSR} 
\begin{figure}[!htbp]
    \centering
    \includegraphics[width=0.9\linewidth]{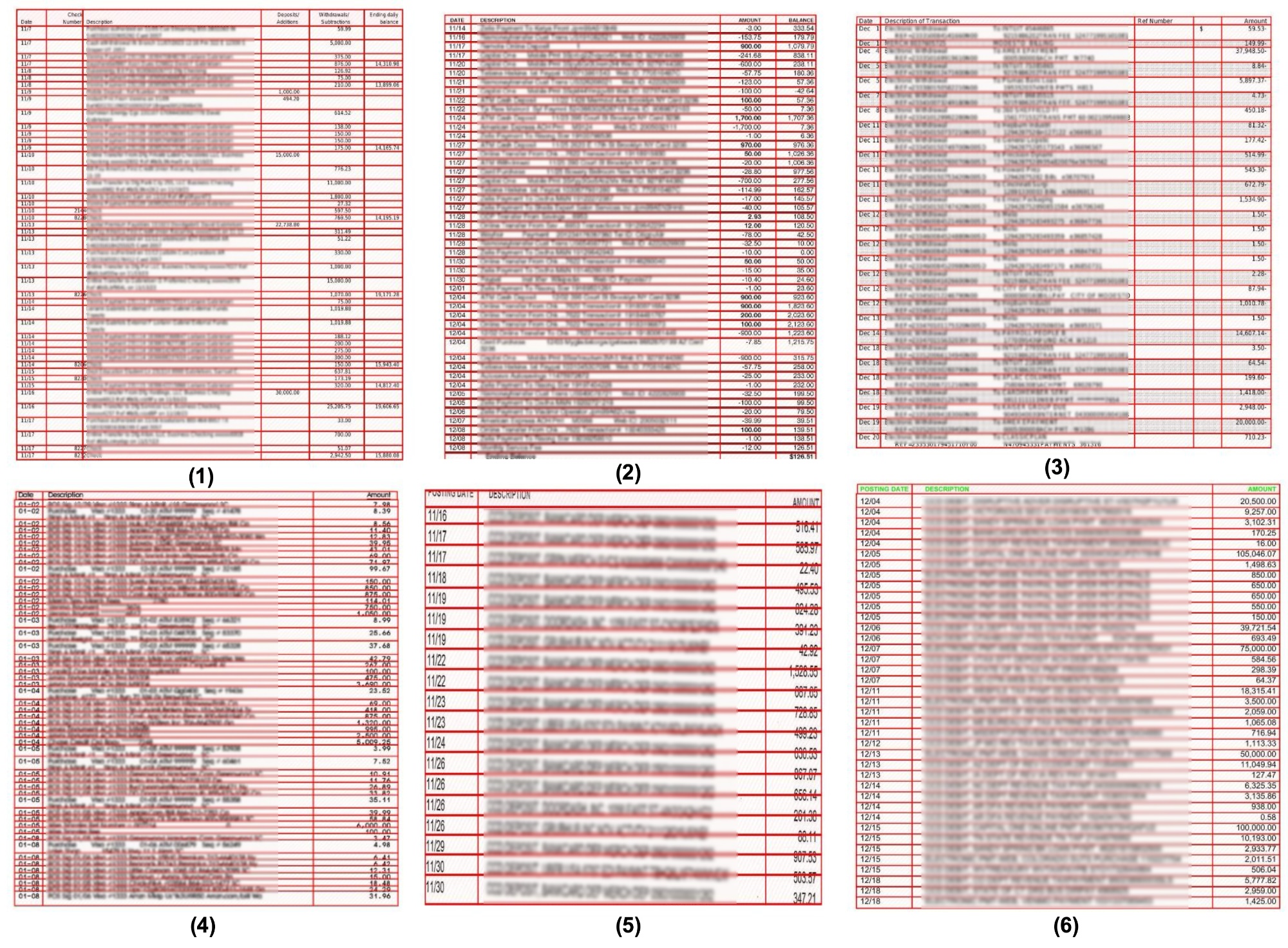}
    \caption{Final cells predicted by TSR on different bank tables.}
    \label{fig:result_ours}
\end{figure}

Final cells predictions on table images from BankTabNet Test Set is shown in Figure \ref{fig:result_ours}. TSR module handles variety of bank templates and table layouts accurately for downstream analysis tasks.
There were few limitation cases of scanned documents (captured through mobile phones etc.) in out of time testing samples where our approach failed because of tilted tables.
Example of one of the table image in shown in sample \textbf{5} of Figure \ref{fig:result_ours}. In this particular image, the bounding boxes’ boundaries predicted by TabSniper is cutting the text which ultimately leads to partial capture of OCR text words. This can be solved by pre-computing deviation angle of document and then rotating the bounding boxes predictions. We will incorporate this feature in future versions of our TabSniper model.
\subsubsection{Qualitative Baseline Comparison}
\begin{figure}
    \centering
    \includegraphics[width=0.9\linewidth]{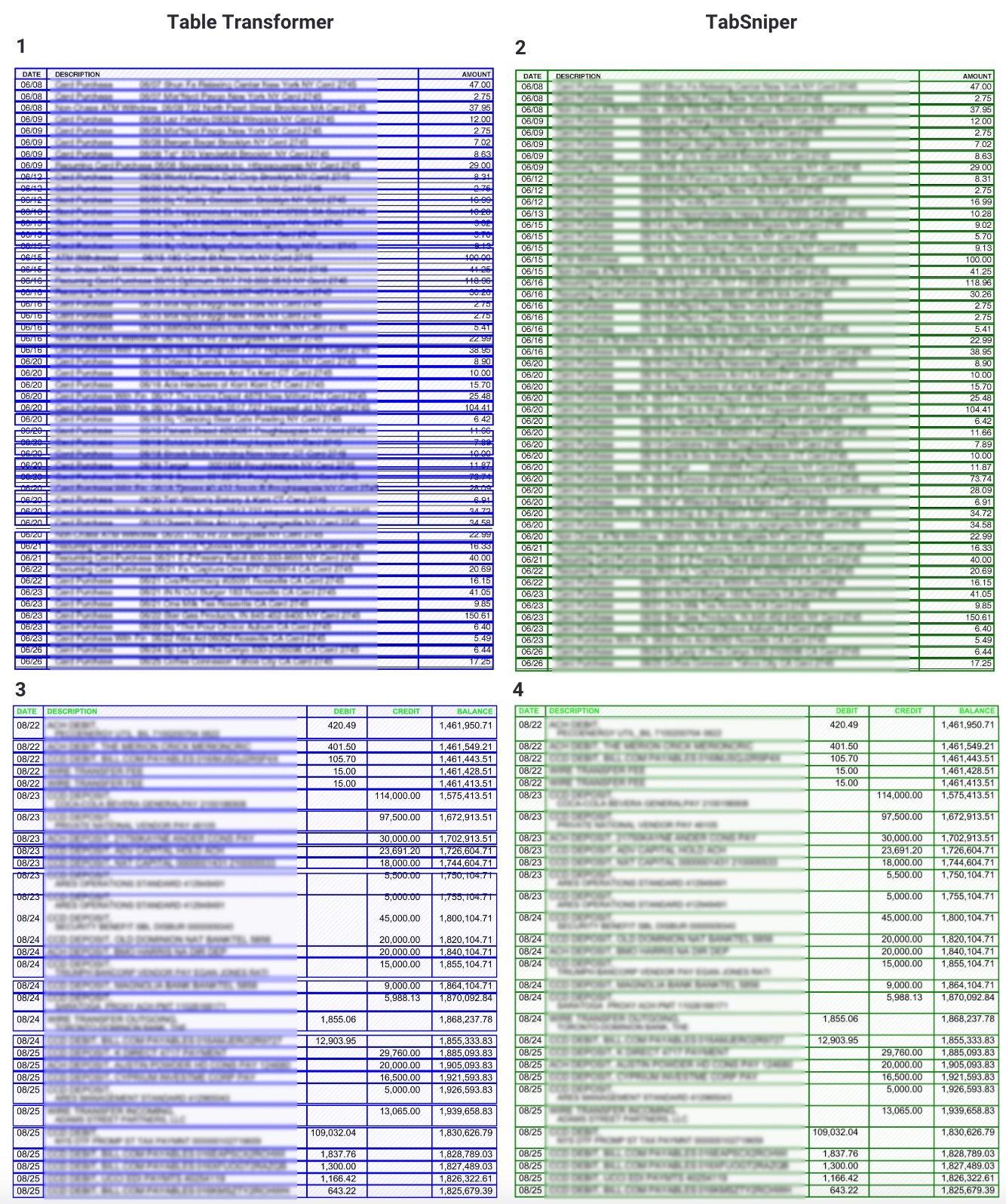}
    \caption{Final table cell bounding boxes predicted by TabSniper(right) in comparison to Table Transformer \cite{smock2021pubtables1m}. Both the models are trained on BankTabNet.}
    \label{fig:result}
\end{figure}

We compare TabSniper with second best baseline Table Transformer \cite{smock2021pubtables1m} in Figure \ref{fig:result}. Both the models comprises of table detection and table structure recognition sub-modules and are trained on BankTabNet. 
TabSniper- TSR detections on table images are better than Table Transformer outputs. With the same base model i.e DETR \cite{carion2020endtoend} in both approaches, our additional feature training with different ablations solve the problem of inaccurate table objects detection. For instance, in table image \textbf{2} of Figure \ref{fig:result}, TabSniper outperforms baseline with long table split-merge approach. The additional row separator info at TSR post-processing is used to refine row predictions. Table row boundaries are cutting text inside table for baseline results, which eventually leads to partial capture of OCR words in final output of table image \textbf{1}.
In table image \textbf{4} of Figure \ref{fig:result}, TabSniper gives precise detection for multi-line rows which were missed by baseline in image \textbf{3} (at table center). TabSniper overcome these issues with innovative data preparation (Section \ref{sec:TSR_dataset}) for model training along with extensive post-processing. Additional penalty terms in Complete IoU loss (refer Section \ref{sec:tsr_post}) helps in bounding boxes alignment. With the specific use-case requirements for capturing every transaction correctly in all the page images of bank statement, TabSniper produce most precise results for bank statement spreading.
\subsubsection{External Datasets:} 
\begin{figure}[!htbp]
    \centering
    \includegraphics[width=0.9\linewidth]{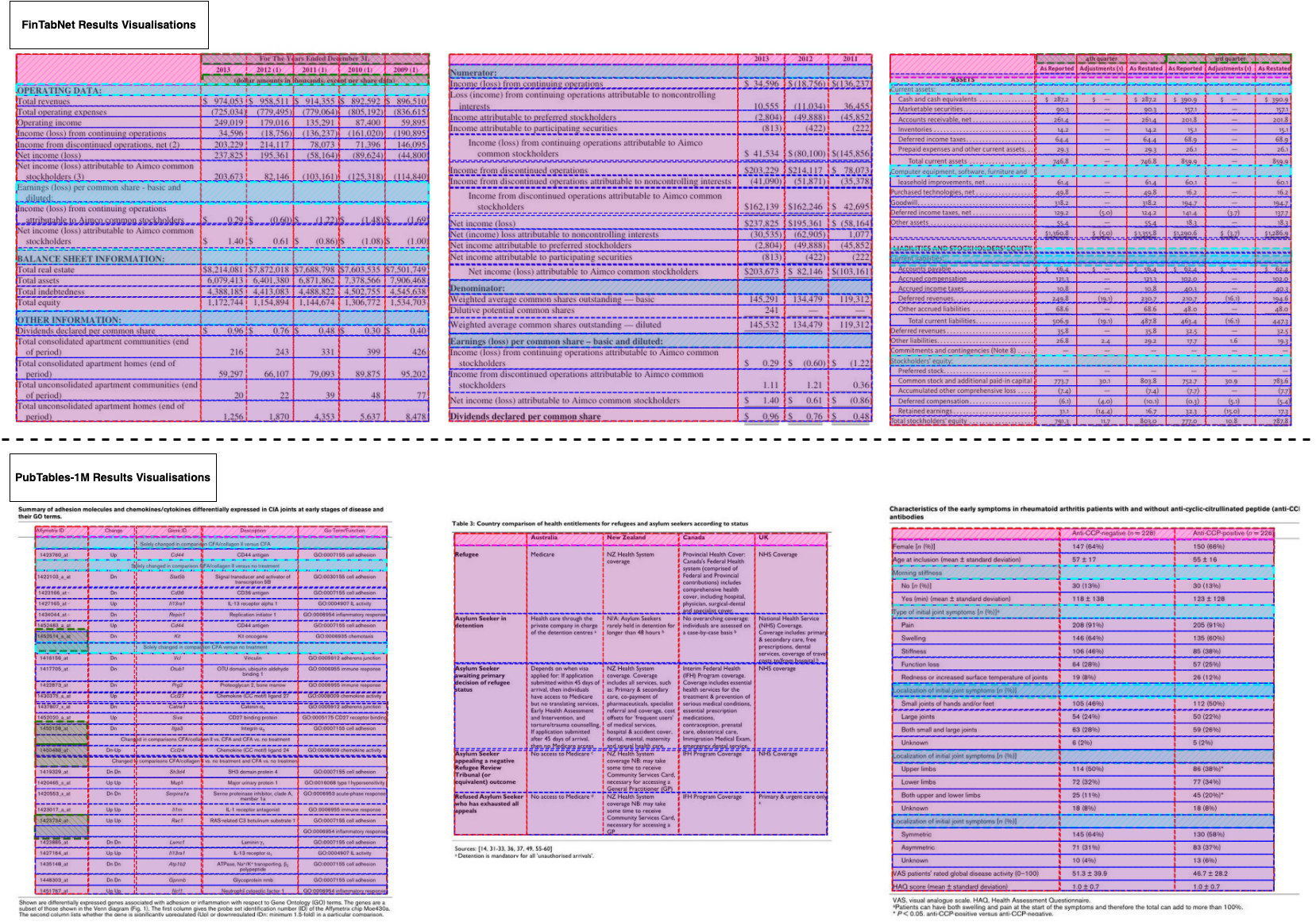}
    \caption{Bounding boxes predicted by TabSniper on external datasets. The colors distinguish region labels. The dotted line separates FinTabNet table images (top) and those
from PubTables-1M (bottom). Note: TabSniper has been trained on respective datasets.}
    \label{fig:result_all}
\end{figure}

Example table images from publicly available table datasets with TabSniper predictions overlaid can be seen in Figure \ref{fig:result_all}. The images above the dotted line are from the FinTabNet\cite{zheng2021global} test dataset. The tables are characterized by ample projected row headers along with dense columns. Despite this, TabSniper provides accurate table objects' detections. Tables images below dotted line are from PubTables-1M\cite{smock2021pubtables1m} test dataset and are characterized by lengthy table text rows along with unwanted text descriptions outside of the table area. TabSniper is trained to handle variety of padding spaces outside table area and detects accurate table bounding boxes along with other table objects.
\vspace{-1em}
\section{Conclusion}
\label{sec:conclusion}
In this paper, we propose TabSniper, a novel approach for automatic bank statement transaction spreading. There are significant contributions from our work presented in paper. (i) Creation of new BankTabNet dataset consisting of table annotations at both page level and transaction level (ii) Table Detection and Categorisation sub module to detect different table categories in bank statements. (iii) Table Structure Recognition module with additional feature model training over DETR to precisely detect every transaction in full table images. Through experimental setup, we demonstrate that TabSniper generates accurate transaction level layout while capturing full transaction description through precise table cell boundaries. TabSniper outperforms strong baselines and ablative variants in quantitative evaluation. Qualitative results shows efficiency of TabSniper to handle densely packed multi-line rows in table and give the most precise table spreading results. 
Finally, we show that TabSniper readily generalizes to other publicly available financial, scientific table datasets. As future works, we plan to handle both regular and irregular tables (rotated, distorted) from other domains like diverse forms and scanned, historical documents.
\bibliographystyle{ACM-Reference-Format}
\bibliography{sample-base}


\begin{thebibliography}{23}


\ifx \showCODEN    \undefined \def \showCODEN     #1{\unskip}     \fi
\ifx \showDOI      \undefined \def \showDOI       #1{#1}\fi
\ifx \showISBNx    \undefined \def \showISBNx     #1{\unskip}     \fi
\ifx \showISBNxiii \undefined \def \showISBNxiii  #1{\unskip}     \fi
\ifx \showISSN     \undefined \def \showISSN      #1{\unskip}     \fi
\ifx \showLCCN     \undefined \def \showLCCN      #1{\unskip}     \fi
\ifx \shownote     \undefined \def \shownote      #1{#1}          \fi
\ifx \showarticletitle \undefined \def \showarticletitle #1{#1}   \fi
\ifx \showURL      \undefined \def \showURL       {\relax}        \fi
\providecommand\bibfield[2]{#2}
\providecommand\bibinfo[2]{#2}
\providecommand\natexlab[1]{#1}
\providecommand\showeprint[2][]{arXiv:#2}

\bibitem[Baviskar et~al\mbox{.}(2021)]%
        {baviskar2021efficient}
\bibfield{author}{\bibinfo{person}{Dipali Baviskar}, \bibinfo{person}{Swati Ahirrao}, \bibinfo{person}{Vidyasagar Potdar}, {and} \bibinfo{person}{Ketan Kotecha}.} \bibinfo{year}{2021}\natexlab{}.
\newblock \showarticletitle{Efficient automated processing of the unstructured documents using artificial intelligence: A systematic literature review and future directions}.
\newblock \bibinfo{journal}{\emph{IEEE Access}}  \bibinfo{volume}{9} (\bibinfo{year}{2021}), \bibinfo{pages}{72894--72936}.
\newblock


\bibitem[Cai and Vasconcelos(2017)]%
        {cai2017cascade}
\bibfield{author}{\bibinfo{person}{Zhaowei Cai} {and} \bibinfo{person}{Nuno Vasconcelos}.} \bibinfo{year}{2017}\natexlab{}.
\newblock \bibinfo{title}{Cascade R-CNN: Delving into High Quality Object Detection}.
\newblock
\newblock
\showeprint[arxiv]{1712.00726}~[cs.CV]


\bibitem[Carion et~al\mbox{.}(2020)]%
        {carion2020endtoend}
\bibfield{author}{\bibinfo{person}{Nicolas Carion}, \bibinfo{person}{Francisco Massa}, \bibinfo{person}{Gabriel Synnaeve}, \bibinfo{person}{Nicolas Usunier}, \bibinfo{person}{Alexander Kirillov}, {and} \bibinfo{person}{Sergey Zagoruyko}.} \bibinfo{year}{2020}\natexlab{}.
\newblock \bibinfo{title}{End-to-End Object Detection with Transformers}.
\newblock
\newblock
\showeprint[arxiv]{2005.12872}~[cs.CV]


\bibitem[Dai et~al\mbox{.}(2021)]%
        {Dai_2021_CVPR}
\bibfield{author}{\bibinfo{person}{Xiyang Dai}, \bibinfo{person}{Yinpeng Chen}, \bibinfo{person}{Bin Xiao}, \bibinfo{person}{Dongdong Chen}, \bibinfo{person}{Mengchen Liu}, \bibinfo{person}{Lu Yuan}, {and} \bibinfo{person}{Lei Zhang}.} \bibinfo{year}{2021}\natexlab{}.
\newblock \showarticletitle{Dynamic Head: Unifying Object Detection Heads With Attentions}. In \bibinfo{booktitle}{\emph{Proceedings of the IEEE/CVF Conference on Computer Vision and Pattern Recognition (CVPR)}}. \bibinfo{pages}{7373--7382}.
\newblock


\bibitem[Gerling and Lessmann(2023)]%
        {gerling2023multimodal}
\bibfield{author}{\bibinfo{person}{Christopher Gerling} {and} \bibinfo{person}{Stefan Lessmann}.} \bibinfo{year}{2023}\natexlab{}.
\newblock \showarticletitle{Multimodal Document Analytics for Banking Process Automation}.
\newblock \bibinfo{journal}{\emph{arXiv preprint arXiv:2307.11845}} (\bibinfo{year}{2023}).
\newblock


\bibitem[Gilani et~al\mbox{.}(2017)]%
        {gilani2017table}
\bibfield{author}{\bibinfo{person}{Azka Gilani}, \bibinfo{person}{Shah~Rukh Qasim}, \bibinfo{person}{Imran Malik}, {and} \bibinfo{person}{Faisal Shafait}.} \bibinfo{year}{2017}\natexlab{}.
\newblock \showarticletitle{Table detection using deep learning}. In \bibinfo{booktitle}{\emph{2017 14th IAPR international conference on document analysis and recognition (ICDAR)}}, Vol.~\bibinfo{volume}{1}. IEEE, \bibinfo{pages}{771--776}.
\newblock


\bibitem[He et~al\mbox{.}(2017)]%
        {8237584}
\bibfield{author}{\bibinfo{person}{Kaiming He}, \bibinfo{person}{Georgia Gkioxari}, \bibinfo{person}{Piotr Dollár}, {and} \bibinfo{person}{Ross Girshick}.} \bibinfo{year}{2017}\natexlab{}.
\newblock \showarticletitle{Mask R-CNN}. In \bibinfo{booktitle}{\emph{2017 IEEE International Conference on Computer Vision (ICCV)}}. \bibinfo{pages}{2980--2988}.
\newblock
\urldef\tempurl%
\url{https://doi.org/10.1109/ICCV.2017.322}
\showDOI{\tempurl}


\bibitem[Lewis and Young(2019)]%
        {lewis2019fad}
\bibfield{author}{\bibinfo{person}{Craig Lewis} {and} \bibinfo{person}{Steven Young}.} \bibinfo{year}{2019}\natexlab{}.
\newblock \showarticletitle{Fad or future? Automated analysis of financial text and its implications for corporate reporting}.
\newblock \bibinfo{journal}{\emph{Accounting and Business Research}} \bibinfo{volume}{49}, \bibinfo{number}{5} (\bibinfo{year}{2019}), \bibinfo{pages}{587--615}.
\newblock


\bibitem[Li et~al\mbox{.}(2022)]%
        {li2022dit}
\bibfield{author}{\bibinfo{person}{Junlong Li}, \bibinfo{person}{Yiheng Xu}, \bibinfo{person}{Tengchao Lv}, \bibinfo{person}{Lei Cui}, \bibinfo{person}{Cha Zhang}, {and} \bibinfo{person}{Furu Wei}.} \bibinfo{year}{2022}\natexlab{}.
\newblock \bibinfo{title}{DiT: Self-supervised Pre-training for Document Image Transformer}.
\newblock
\newblock
\showeprint[arxiv]{2203.02378}~[cs.CV]


\bibitem[Manning et~al\mbox{.}(2008)]%
        {manning-2008}
\bibfield{author}{\bibinfo{person}{Christopher~D. Manning}, \bibinfo{person}{Prabhakar Raghavan}, {and} \bibinfo{person}{Hinrich Schütze}.} \bibinfo{year}{2008}\natexlab{}.
\newblock \bibinfo{booktitle}{\emph{{Introduction to Information Retrieval}}}.
\newblock \bibinfo{publisher}{Cambridge University Press}, \bibinfo{pages}{234--265}.
\newblock


\bibitem[Oral et~al\mbox{.}(2019)]%
        {oral2019extracting}
\bibfield{author}{\bibinfo{person}{Berke Oral}, \bibinfo{person}{Erdem Emekligil}, \bibinfo{person}{Se{\c{c}}il Arslan}, {and} \bibinfo{person}{G{\"u}l{\c{s}}en Eryi{\u{g}}it}.} \bibinfo{year}{2019}\natexlab{}.
\newblock \showarticletitle{Extracting complex relations from banking documents}. In \bibinfo{booktitle}{\emph{Proceedings of the Second Workshop on Economics and Natural Language Processing}}. \bibinfo{pages}{1--9}.
\newblock


\bibitem[Paliwal et~al\mbox{.}(2019)]%
        {paliwal2019tablenet}
\bibfield{author}{\bibinfo{person}{Shubham~Singh Paliwal}, \bibinfo{person}{D Vishwanath}, \bibinfo{person}{Rohit Rahul}, \bibinfo{person}{Monika Sharma}, {and} \bibinfo{person}{Lovekesh Vig}.} \bibinfo{year}{2019}\natexlab{}.
\newblock \showarticletitle{Tablenet: Deep learning model for end-to-end table detection and tabular data extraction from scanned document images}. In \bibinfo{booktitle}{\emph{2019 International Conference on Document Analysis and Recognition (ICDAR)}}. IEEE, \bibinfo{pages}{128--133}.
\newblock


\bibitem[Prasad et~al\mbox{.}(2020)]%
        {prasad2020cascadetabnet}
\bibfield{author}{\bibinfo{person}{Devashish Prasad}, \bibinfo{person}{Ayan Gadpal}, \bibinfo{person}{Kshitij Kapadni}, \bibinfo{person}{Manish Visave}, {and} \bibinfo{person}{Kavita Sultanpure}.} \bibinfo{year}{2020}\natexlab{}.
\newblock \showarticletitle{CascadeTabNet: An approach for end to end table detection and structure recognition from image-based documents}. In \bibinfo{booktitle}{\emph{Proceedings of the IEEE/CVF conference on computer vision and pattern recognition workshops}}. \bibinfo{pages}{572--573}.
\newblock


\bibitem[Qiao et~al\mbox{.}(2022)]%
        {qiao2022lgpma}
\bibfield{author}{\bibinfo{person}{Liang Qiao}, \bibinfo{person}{Zaisheng Li}, \bibinfo{person}{Zhanzhan Cheng}, \bibinfo{person}{Peng Zhang}, \bibinfo{person}{Shiliang Pu}, \bibinfo{person}{Yi Niu}, \bibinfo{person}{Wenqi Ren}, \bibinfo{person}{Wenming Tan}, {and} \bibinfo{person}{Fei Wu}.} \bibinfo{year}{2022}\natexlab{}.
\newblock \bibinfo{title}{LGPMA: Complicated Table Structure Recognition with Local and Global Pyramid Mask Alignment}.
\newblock
\newblock
\showeprint[arxiv]{2105.06224}~[cs.CV]


\bibitem[Raja et~al\mbox{.}(2020)]%
        {raja2020table}
\bibfield{author}{\bibinfo{person}{Sachin Raja}, \bibinfo{person}{Ajoy Mondal}, {and} \bibinfo{person}{C.~V. Jawahar}.} \bibinfo{year}{2020}\natexlab{}.
\newblock \bibinfo{title}{Table Structure Recognition using Top-Down and Bottom-Up Cues}.
\newblock
\newblock
\showeprint[arxiv]{2010.04565}~[cs.CV]


\bibitem[Ren et~al\mbox{.}(2017)]%
        {7485869}
\bibfield{author}{\bibinfo{person}{Shaoqing Ren}, \bibinfo{person}{Kaiming He}, \bibinfo{person}{Ross Girshick}, {and} \bibinfo{person}{Jian Sun}.} \bibinfo{year}{2017}\natexlab{}.
\newblock \showarticletitle{Faster R-CNN: Towards Real-Time Object Detection with Region Proposal Networks}.
\newblock \bibinfo{journal}{\emph{IEEE Transactions on Pattern Analysis and Machine Intelligence}} \bibinfo{volume}{39}, \bibinfo{number}{6} (\bibinfo{year}{2017}), \bibinfo{pages}{1137--1149}.
\newblock
\urldef\tempurl%
\url{https://doi.org/10.1109/TPAMI.2016.2577031}
\showDOI{\tempurl}


\bibitem[Schreiber et~al\mbox{.}(2017)]%
        {schreiber2017deepdesrt}
\bibfield{author}{\bibinfo{person}{Sebastian Schreiber}, \bibinfo{person}{Stefan Agne}, \bibinfo{person}{Ivo Wolf}, \bibinfo{person}{Andreas Dengel}, {and} \bibinfo{person}{Sheraz Ahmed}.} \bibinfo{year}{2017}\natexlab{}.
\newblock \showarticletitle{Deepdesrt: Deep learning for detection and structure recognition of tables in document images}. In \bibinfo{booktitle}{\emph{2017 14th IAPR international conference on document analysis and recognition (ICDAR)}}, Vol.~\bibinfo{volume}{1}. IEEE, \bibinfo{pages}{1162--1167}.
\newblock


\bibitem[Smock et~al\mbox{.}(2021)]%
        {smock2021pubtables1m}
\bibfield{author}{\bibinfo{person}{Brandon Smock}, \bibinfo{person}{Rohith Pesala}, {and} \bibinfo{person}{Robin Abraham}.} \bibinfo{year}{2021}\natexlab{}.
\newblock \bibinfo{title}{PubTables-1M: Towards comprehensive table extraction from unstructured documents}.
\newblock
\newblock
\showeprint[arxiv]{2110.00061}~[cs.LG]


\bibitem[Sun et~al\mbox{.}(2021)]%
        {sun2021sparse}
\bibfield{author}{\bibinfo{person}{Peize Sun}, \bibinfo{person}{Rufeng Zhang}, \bibinfo{person}{Yi Jiang}, \bibinfo{person}{Tao Kong}, \bibinfo{person}{Chenfeng Xu}, \bibinfo{person}{Wei Zhan}, \bibinfo{person}{Masayoshi Tomizuka}, \bibinfo{person}{Lei Li}, \bibinfo{person}{Zehuan Yuan}, \bibinfo{person}{Changhu Wang}, {et~al\mbox{.}}} \bibinfo{year}{2021}\natexlab{}.
\newblock \showarticletitle{Sparse r-cnn: End-to-end object detection with learnable proposals}. In \bibinfo{booktitle}{\emph{Proceedings of the IEEE/CVF conference on computer vision and pattern recognition}}. \bibinfo{pages}{14454--14463}.
\newblock


\bibitem[Tschirschwitz et~al\mbox{.}(2022)]%
        {kalpha}
\bibfield{author}{\bibinfo{person}{David Tschirschwitz}, \bibinfo{person}{Franziska Klemstein}, \bibinfo{person}{Benno Stein}, {and} \bibinfo{person}{Volker Rodehorst}.} \bibinfo{year}{2022}\natexlab{}.
\newblock \showarticletitle{A Dataset for Analysing Complex Document Layouts in the Digital Humanities and its Evaluation with Krippendorff ’s Alpha}. In \bibinfo{booktitle}{\emph{Pattern Recognition}}. \bibinfo{publisher}{Springer International Publishing}, \bibinfo{address}{Cham}, \bibinfo{pages}{354--374}.
\newblock


\bibitem[Wu et~al\mbox{.}(2019)]%
        {wu2019detectron2}
\bibfield{author}{\bibinfo{person}{Yuxin Wu}, \bibinfo{person}{Alexander Kirillov}, \bibinfo{person}{Francisco Massa}, \bibinfo{person}{Wan-Yen Lo}, {and} \bibinfo{person}{Ross Girshick}.} \bibinfo{year}{2019}\natexlab{}.
\newblock \bibinfo{title}{Detectron2}.
\newblock \bibinfo{howpublished}{\url{https://github.com/facebookresearch/detectron2}}.
\newblock


\bibitem[Zhang et~al\mbox{.}(2023)]%
        {zhang2023dense}
\bibfield{author}{\bibinfo{person}{Shilong Zhang}, \bibinfo{person}{Xinjiang Wang}, \bibinfo{person}{Jiaqi Wang}, \bibinfo{person}{Jiangmiao Pang}, \bibinfo{person}{Chengqi Lyu}, \bibinfo{person}{Wenwei Zhang}, \bibinfo{person}{Ping Luo}, {and} \bibinfo{person}{Kai Chen}.} \bibinfo{year}{2023}\natexlab{}.
\newblock \showarticletitle{Dense Distinct Query for End-to-End Object Detection}. In \bibinfo{booktitle}{\emph{Proceedings of the IEEE/CVF Conference on Computer Vision and Pattern Recognition}}. \bibinfo{pages}{7329--7338}.
\newblock


\bibitem[Zheng et~al\mbox{.}(2021)]%
        {zheng2021global}
\bibfield{author}{\bibinfo{person}{Xinyi Zheng}, \bibinfo{person}{Douglas Burdick}, \bibinfo{person}{Lucian Popa}, \bibinfo{person}{Xu Zhong}, {and} \bibinfo{person}{Nancy Xin~Ru Wang}.} \bibinfo{year}{2021}\natexlab{}.
\newblock \showarticletitle{Global table extractor (gte): A framework for joint table identification and cell structure recognition using visual context}. In \bibinfo{booktitle}{\emph{Proceedings of the IEEE/CVF winter conference on applications of computer vision}}. \bibinfo{pages}{697--706}.
\newblock


\end{thebibliography}

\end{document}